\documentclass[letterpaper, 10 pt, conference]{ieeeconf}  

\IEEEoverridecommandlockouts                              

\usepackage{graphics} 
\usepackage{epsfig} 
\usepackage{times} 
\usepackage{amsmath} 
\usepackage{amssymb}  
\usepackage[table]{xcolor}
\usepackage{mathtools,commath}
\usepackage{graphicx,import}
\usepackage{verbatim}
\usepackage{color, soul}
\usepackage{lipsum}
\usepackage{verbatim}
\usepackage{algorithm}
\usepackage[noend]{algpseudocode}
\usepackage{multicol}
\usepackage{subfigure}
\usepackage{tikz}
\usepackage{booktabs}
\usepackage[utf8]{inputenc}
\usepackage[english]{babel}
\usepackage{todonotes}
\usepackage{diagbox}
\usepackage{color, colortbl}
\usepackage{microtype}
\usepackage{siunitx} 
\usepackage{gensymb} 
\usepackage{cite}
\addto\captionsenglish{}

\definecolor{Gray}{gray}{0.9}
\newcolumntype{g}{>{\columncolor{Gray}}c}

\usepackage{xurl} 
\usepackage[breaklinks, citecolor=black]{hyperref}
\usepackage{multirow}

\title{\LARGE \bf
Centroidal Aerodynamic Modeling and Control  \\ of Flying Multibody Robots 
}

\author{Tong Hui$^{1,2,\textbf{*}}$, Antonello Paolino$^{1,4,\textbf{*}}$, Gabriele Nava$^{1}$, Giuseppe L'Erario$^{1,3}$, Fabio Di Natale$^{1}$,\\ Fabio Bergonti$^{1,3}$, Francesco Braghin$^{2}$ and Daniele Pucci$^{1,3}$

\thanks{$^\textbf{*}$ \textbf{The two authors equally contributed to the paper.}}%
\thanks{$^{1}$ Artificial and Mechanical Intelligence, Istituto Italiano di Tecnologia, Genova, Italy {\tt\small firstname.surname@iit.it}}%
\thanks{$^{2}$ Department of Mechanical Engineering, Politecnico di Milano, Milan, Italy {\tt\small francesco.braghin@polimi.it} {\tt\small tong.hui@mail.polimi.it}}%
\thanks{$^{3}$ School of Computer Science, Univ. of Manchester, Manchester, U.K.}
\thanks{$^{4}$ Department of Industrial Engineering, Universit$\grave{a}$ degli Studi di Napoli Federico II, Naples, Italy {\tt\small anton.paolino@studenti.unina.it}} 
}

\begin{document}

\maketitle
\thispagestyle{empty}
\pagestyle{empty}

\begin{abstract}

This paper presents a modeling and control framework for multibody flying robots  subject to non-negligible aerodynamic forces acting on the centroidal dynamics.
First, aerodynamic forces are calculated during robot flight in different operating conditions by means of Computational Fluid Dynamics (CFD) analysis.
Then, analytical models of the aerodynamics coefficients are generated from the dataset collected with CFD analysis. The obtained simplified aerodynamic model is also used to improve the flying robot control design. We present two control strategies: compensating for the aerodynamic effects via feedback linearization and enforcing the controller robustness with gain-scheduling. Simulation results on the jet-powered humanoid robot iRonCub validate the proposed approach.

\end{abstract}


\section{Introduction}

Flying vehicles remain an active research domain for the Robotics community after decades of studies in the subject. 
For instance, inspection via \emph{passive} Vertical Take Off and Landing (VTOL) systems and grasping and manipulation using \emph{active} flying vehicles represent   research problems that still call for new tools and methods \cite{c8,c9,c10}. Often, the control of \emph{small} aircraft neglects the aerodynamic effects acting on the body, which is a barrier for improving the vehicle performance and flight envelope \cite{c11,c13}. This paper proposes an approach that considers advanced aerodynamic modelling techniques for the conception of simplified  models then used on-line for the control of flying multibody robots.

\begin{figure}[t]
	\centering
	\includegraphics[width=0.7\columnwidth]{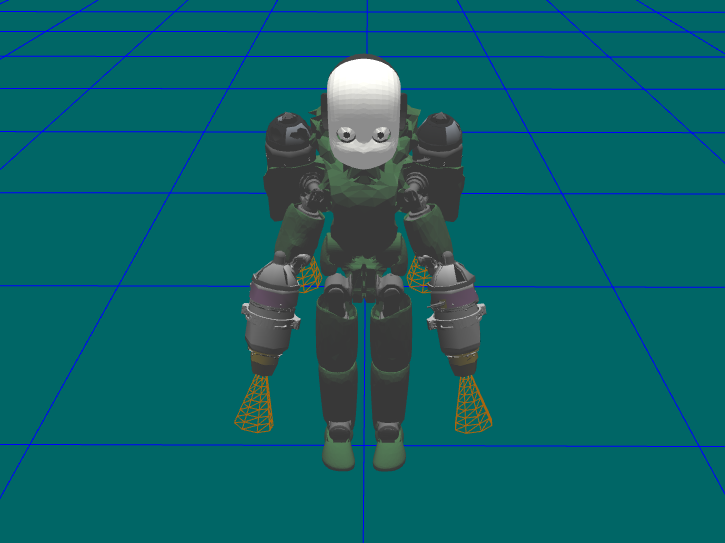}
	\caption{The jet-powered humanoid robot iRonCub.}
	\label{fig:iron_cad}
\end{figure}

The problem of flying vehicle control is especially challenging for the so-called multimodal robots, which attempt at combining 
terrestrial and aerial locomotion in a single robotic platform.
Hexapod-quadrotors, insect biobots, and humanoid robots with thrusters are examples of platforms that combine different degrees of locomotion, thus requiring advanced control and modeling methods \cite{pitonyak2017hexapod,kalantari2013hytaq,bozkurt2009insectbio,LeoCaltech,Jet-HR1}. In this category of platforms there is iRonCub, a prototype of flying humanoid robot currently developed at the Italian Institute of Technology 
~\cite{c1,c2,Giuse,9622189}.
~As shown in Fig.~\ref{fig:iron_cad}, four jet engines located on the robot arms and chest are used to lift from the ground a reworked version of the longstanding iCub platform \cite{Bartolozzi2017iCubTN}.


A common assumption for high DoF VTOL systems is that the vehicle operates at relatively \emph{small speed}, often in indoor environments \cite{c8,c9,c10}, and  consequently the aerodynamic forces acting on the system are usually negligible and not considered in modeling and control framework \cite{c11,c28}. When aerodynamic effects are important enough to affect system stability during flight, or high precision is required for performing a task, different modeling and control strategies could be adopted, e.g. to estimate aerodynamic force by designing a momentum-based observer or other disturbances estimation strategies \cite{c17,c16} and to compensate estimated aerodynamic force by means of adaptive control strategies \cite{c14,c15}. However, the absence of an explicit model of the aerodynamic forces could impair the effectiveness of these methods for exploiting physical properties that are beneficial for flight.\looseness=-1

%

High-speed flight can benefit from the presence of lift forces alleviating the effort made by the propellers for gravity compensation. The VTOL dynamical model, in this case, explicitly includes aerodynamic forces \cite{c3,c18,KAI2019108491}. Few examples include the analysis of blade flapping on quadrotors, modeling first-order aerodynamic effects and evaluating the effect of drag force on thrust power under hovering, \cite{c7}\cite{c20}\cite{c19}.
%
%
%
However, the shape of aerial vehicles has a significant role in modeling aerodynamics effects. It is difficult to design an effective aerodynamic model in case of complex-shaped, high DoF aerial systems \cite{c6,c18}.

%

Different techniques have been developed to properly estimate aerodynamic forces for complex-shaped systems, e.g. wind tunnel test and Computational Fluid Dynamics (CFD) analysis. However, to estimate the aerodynamic forces for an entire set of robot operating conditions, i.e. \emph{flight envelope}, it requires to collect a very large number of data. An alternative is to design simpler analytical models via limited dataset provided by wind tunnel tests or CFD simulations. Many of these analytical models assume that the robot body is axisymmetric \cite{c3,c23,c25}.  

This paper proposes a framework to include aerodynamic forces in the modeling and control design of high DoF VTOL systems, which in the case under study are represented by the flying humanoid robot iRonCub. 
%
At first, a series of CFD simulations is performed on a simplified robot model, in order to generate a dataset of the aerodynamic forces acting on the robot for a given flight envelope. Secondly, a model for a non-axisymmetric bluff body is designed to fit the collected dataset and estimate the aerodynamic characteristics in between the flight envelope. 
Finally, the model is used to improve the iRonCub flight-control strategy developed in our previous works \cite{c1}\cite{c2} by compensation of the aerodynamic forces via \emph{feedback linearization} and improvement of the controller robustness with \emph{gain scheduling}.


The paper is organized as follows. Sec.~\ref{background} recalls notation, system modeling and fundamentals of aerodynamic modeling. In Sec.~\ref{methods}, we describe the CFD analysis on a simplified iRonCub model and the identification of the aerodynamic characteristics. Sec.~\ref{sec:control} presents the improvements of iRonCub flight controller to handle aerodynamic effects. Simulation results are displayed in Sec.~\ref{simulation}. Sec.~\ref{conclusion} concludes the paper and points out future directions.

\section{Background}
\label{background}

\subsection{Notation}
\begin{itemize}
	\item $\mathcal{I}=\{O;\boldsymbol{X},\boldsymbol{Y},\boldsymbol{Z}\}$ denotes an inertial frame, composed of a point (origin) and an orientation w.r.t. which the vehicle’s absolute pose is measured.
	
	\item $\mathcal{B}$ denotes the \emph{base frame}, i.e. a frame rigidly attached to the robot \emph{base link};
	
	\item $\mathcal{G}[\mathcal{I}]$ is a frame with the origin at the robot center of mass and the same orientation of the inertial frame;
	
	\item ${}^{\mathcal{I}}o_{\mathcal{G}} \in \mathbb{R}^3$ is the center of mass (CoM) position of the robot expressed in the inertial frame. ${}^{\mathcal{I}}\boldsymbol{v}_{CoM} \in \mathbb{R}^3$ denotes the linear velocity of the robot CoM w.r.t. the inertial frame origin expressed in the inertial frame;
	
	\item ${}^{\mathcal{I}}\boldsymbol{v}_a \in \mathbb{R}^3$ indicates the relative velocity between the wind and the robot CoM expressed in the inertial frame, i.e. ${}^{\mathcal{I}}\boldsymbol{v}_a={}^{\mathcal{I}}\boldsymbol{v}_{CoM}-{}^{\mathcal{I}}\boldsymbol{v}_w$, with ${}^{\mathcal{I}}\boldsymbol{v}_w \in \mathbb{R}^3$ the absolute wind velocity\footnote{In the remainder of the paper, we omit the superscript $\mathcal{I}$ unless it is necessary to explicitly write it for the sake of clarity, and we write ${}^{\mathcal{I}}\boldsymbol{v}_a=\boldsymbol{v}_a$.};
	
	
	\item $\mathcal{D}=\{{}^{\mathcal{I}}o_{\mathcal{G}};\boldsymbol{i},\boldsymbol{j},\boldsymbol{k}\}$ represents the \emph{body frame}. The $\boldsymbol{k}$ axis direction is parallel to the chest cylinder principal axis, pointing towards the legs. The $\boldsymbol{i}$ axis is normal to the robot symmetry plane, pointing towards left arm. The $\boldsymbol{j}$ axis is defined by the right-hand rule. 
	
	
	
	
\end{itemize}

\subsection{Robot Modeling}
\begin{figure}[t]
	\centering
	\includegraphics[width=0.8\columnwidth]{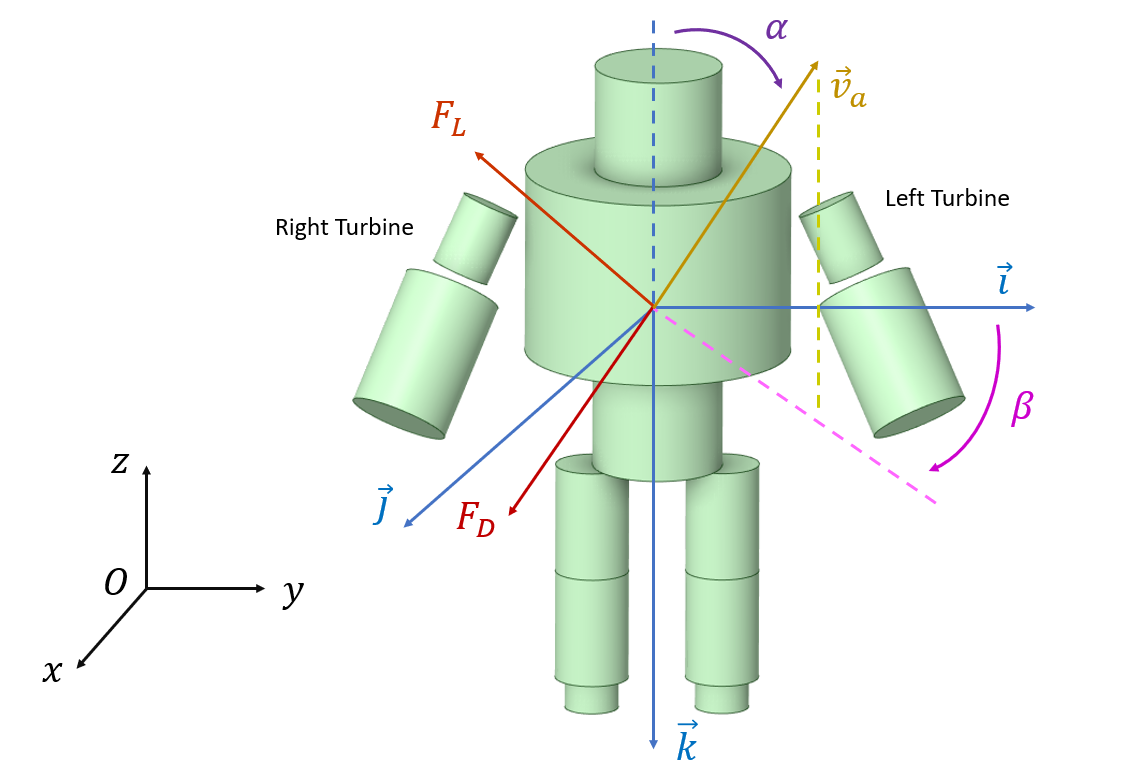}
	\caption{Notation representation: inertial frame, body frame $\mathcal{D}$ (in blue), relative velocity $\boldsymbol{v}_a$ (in yellow) and $(\alpha,\beta)$ angles.} \label{fig:notation} 
\end{figure}

The flying humanoid robot is modeled as a multi-body, \emph{floating-base} system composed of $n+1$ rigid bodies (\emph{links}). Each body is connected to the others by means of 1-DoF joints, for a total of $n$ DoF. The set of robot configurations belongs to a space defined as: $\mathbb{Q} \in \mathbb{R}^3 \times SO(3) \times \mathbb{R}^n$. The triplet $q=( {} ^{\mathcal{I}}o_{\mathcal{B}} , {} ^{\mathcal{I}}R_{\mathcal{B}} , s) \in \mathbb{Q}$ is composed by $( {} ^{\mathcal{I}}o_{\mathcal{B}} , {} ^{\mathcal{I}}R_{\mathcal{B}})$, which represent the position and orientation of the base frame $\mathcal{B}$ w.r.t. the inertial frame, and by the robot internal shape $s \in \mathbb{R}^n$ represented by the joints angles. The system velocities belong to the space $\mathbb{V} \in \mathbb{R}^3 \times \mathbb{R}^3 \times \mathbb{R}^n$. An element of $\mathbb{V}$ is given by $v=(\nu_{\mathcal{B}},\dot{s})$ where $\nu_{\mathcal{B}}=({}^{\mathcal{I}} v_{\mathcal{B}},{}^{\mathcal{I}} \omega_{\mathcal{B}})$ is the linear and angular velocity of the base frame w.r.t. the inertial frame, while $\dot{s}$ are the joints velocities.

The floating-base system equations of motions can be derived following the Euler-Poincar$\acute{e}$ formalism \cite{c4}:
\begin{equation}
	\label{eq_motion}
	M(q)\dot{v}+C(q,v)v+G(q)=
	\begin{bmatrix}
		0_6\\
		\tau
	\end{bmatrix}
	+\sum_{k=1}^{n_c}J_k^{\top} F_k ,
\end{equation}
where $M$, $C \in \mathbb{R}^{n+6\times n+6}$ are the mass and Coriolis matrices, $G \in \mathbb{R}^{n+6}$ is the gravity vector, $\tau \in \mathbb{R}^n$ are the internal actuation torques and $F_k$ is the $k^{th}$ external force acting on the system. In the case under study, we assume that the forces acting on the robot during flight are the aerodynamic forces and the thrust forces generated by the jet-engines. Here, we consider in the model only the thrust generated by the jet-engines, while how to add the aerodynamic forces in Eq.~(\ref{eq_motion}) will be treated later on in the paper. Hence, each $F_k \in \mathbb{R}^3$ represents the thrust force applied on the robot by the $k^{th}$ jet, and we rewrite $F_k={}^\mathcal{I}l_k T_k$, where ${}^\mathcal{I}l_k \in \mathbb{R}^3$ is the direction of thrust force, while $T_k \in \mathbb{R}$ is the thrust intensity. The jacobian $J_k(q)$ is the map between the system velocity $v$ and the linear velocity ${}^\mathcal{I} v_k$ of the $k^{th}$ thrust application point. We can then define the thrusts vector $T:=(T_1, T_2, T_3, T_4)^{\top}$ and rewrite $\sum^{n_c}_{k=1}J_k^{\top}F_k = f(q,T)$.

\subsection{Centroidal dynamics}\label{sec:intro_momentum}

We introduce ${}^{\mathcal{G}[\mathcal{I}]} \boldsymbol{h} \in \mathbb{R}^6$ to denote the robot total momentum expressed w.r.t. ${}^{\mathcal{G}[\mathcal{I}]}$, namely:
\begin{equation}
	{}^{\mathcal{G}[\mathcal{I}]} \boldsymbol{h} = \begin{bmatrix}
		{}_{{}^{\mathcal{G}[\mathcal{I}]}} \boldsymbol{h}^p \\
		{}_{{}^{\mathcal{G}[\mathcal{I}]}} \boldsymbol{h}^\omega
	\end{bmatrix},
\end{equation}
where ${}^{\mathcal{G}[\mathcal{I}]} \boldsymbol{h}^p \in \mathbb{R}^3$ and ${}^{\mathcal{G}[\mathcal{I}]} \boldsymbol{h}^\omega \in \mathbb{R}^3$ are the linear and angular momentum, respectively. In addition, the following holds:
${}^{\mathcal{I}}\boldsymbol{v}_{CoM}  = \frac{1}{m}{} {{}^{\mathcal{G}[\mathcal{I}]}} \boldsymbol{h}^p.$
The robot total momentum corresponds to the summation of all the links momenta ${}_B\boldsymbol{h}^i$, measured in the base frame and projected on ${\mathcal{G}[\mathcal{I}]}$:
${}^{\mathcal{G}[\mathcal{I}]}\boldsymbol{h} = \sum_i {}_{{}^{\mathcal{G}[\mathcal{I}]}} \boldsymbol{X}^B {}_B\boldsymbol{h}^i,$
with ${}_{{}^{\mathcal{G}[\mathcal{I}]}} \boldsymbol{X}^B \in \mathbb{R}^{6\times6}$ the \emph{adjoint matrix}  transforming a wrench expressed in $B$ into one expressed in ${\mathcal{G}[\mathcal{I}]}$. 
Then: 
\begin{equation}\label{eq:momentumExpanded}
	{}^{\mathcal{G}[\mathcal{I}]} \boldsymbol{h} = {}_{{}^{\mathcal{G}[\mathcal{I}]}} \boldsymbol{X}^B \sum_i {}_{B}\boldsymbol{X}^i \boldsymbol{I}_i {}^i \boldsymbol{V}_{A,i},
\end{equation}
with $\boldsymbol{I}_i \in \mathbb{R}^{6\times 6}$ being the (constant) link inertia expressed in link frame. Hence, one obtains:
${}^{\mathcal{G}[\mathcal{I}]} \boldsymbol{h} = \boldsymbol{J}_\text{CMM}\boldsymbol{v},$
where $\boldsymbol{J}_\text{CMM} \in \mathbb{R}^{6\times n}$ is the \emph{Centroidal Momentum Matrix} (CMM) \cite{orin08}.
The rate of change of the centroidal momentum balances the external wrenches applied to the robot, i.e.:
\begin{equation}\label{eq:centroidal_momentum_dynamics}
	\begin{split}
		{}^{\mathcal{G}[\mathcal{I}]}  \dot{\boldsymbol{h}} &= \sum_{k = 1}^{n_c} {}_{{}^{\mathcal{G}[\mathcal{I}]}}\boldsymbol{X}^k {}_k\textbf{f} + m \bar{\boldsymbol{g}}, \\
		&= \sum_{k = 1}^{n_c} \begin{bmatrix}
			{}^{\mathcal{I}}R_k & 0_{3\times 3} \\
			({}^{\mathcal{I}}o_k - x)^\wedge\,{}^{\mathcal{I}}R_k & {}^{\mathcal{I}}R_k
		\end{bmatrix} {}_k\textbf{f} + m \bar{\boldsymbol{g}}, 
	\end{split}
\end{equation}
where ${}_{{}^{\mathcal{G}[\mathcal{I}]}}\boldsymbol{X}^k \in \mathbb{R}^{6 \times 6}$ is the adjoint matrix that transforms an external wrench from the application frame (located in ${}^{\mathcal{I}}o_k$ with orientation ${}^{\mathcal{I}}R_k$) to ${\mathcal{G}[\mathcal{I}]}$. Finally, $\bar{\boldsymbol{g}}$
is the 6D gravity acceleration vector.

In the sequel, we consider the \emph{overall} aerodynamic effects that can be described as an external force $\boldsymbol{F_a} \in \mathbb{R}^3$ applied at the robot center of mass, thus acting on both  \eqref{eq:centroidal_momentum_dynamics} and \eqref{eq_motion}.

\section{Computational Fluid Dynamics for Aerodynamic Force Modeling}
\label{methods}
The objective of this section is to build models for the aerodynamic force $\boldsymbol{F_a}$ to be then considered in \eqref{eq:centroidal_momentum_dynamics} and \eqref{eq_motion}. In general, the aerodynamic force is hard to predict. It depends on the robot relative velocity, joint angles, and robot base orientation, thus rendering the overall task hard to solve. For this reason, this section performs \emph{Computational Fluid Dynamics} (CFD) simulations considering the robot as a \emph{single rigid body}, namely, the robot is kept with a specific joint configuration $\bar{s}$, and the following assumption is made.

\textbf{Assumption 1}: Assume a constant robot base and wind velocity. Given a robot joint configuration $\bar{s}$,
the aerodynamic force is constant 
when $s$ belongs to the neighborhood of $\bar{s}$.

The assumption above will be enforced by the design of flight controllers that keep the robot joints close to a specific configuration during flight.

\subsection{CFD simplified robot model}
\label{sec:cfdModel}


Considering \textbf{Assumption 1}, CFD simulations are all performed with fixed robot initial joint positions, which is supposed to been optimized given a flight metric (e.g. minimisation of thrust forces).
Furthermore, a complex-shaped geometry as the iRonCub one can cause difficulties in the meshing phase of CFD simulations, because of errors caused by interference among complex parts. Therefore, 
we approximate the geometry of iRonCub representing the major robot links as simple cylinders (see Fig.~\ref{fig:notation}). The model allows to obtain high-quality meshes with a relatively low number of elements, which in turn translates in a reduction of the simulation time.

\subsection{Aerodynamic components for CFD analysis}
\label{sec:cfd} 



\begin{figure}[t]
	\centering
	\includegraphics[width=60mm]{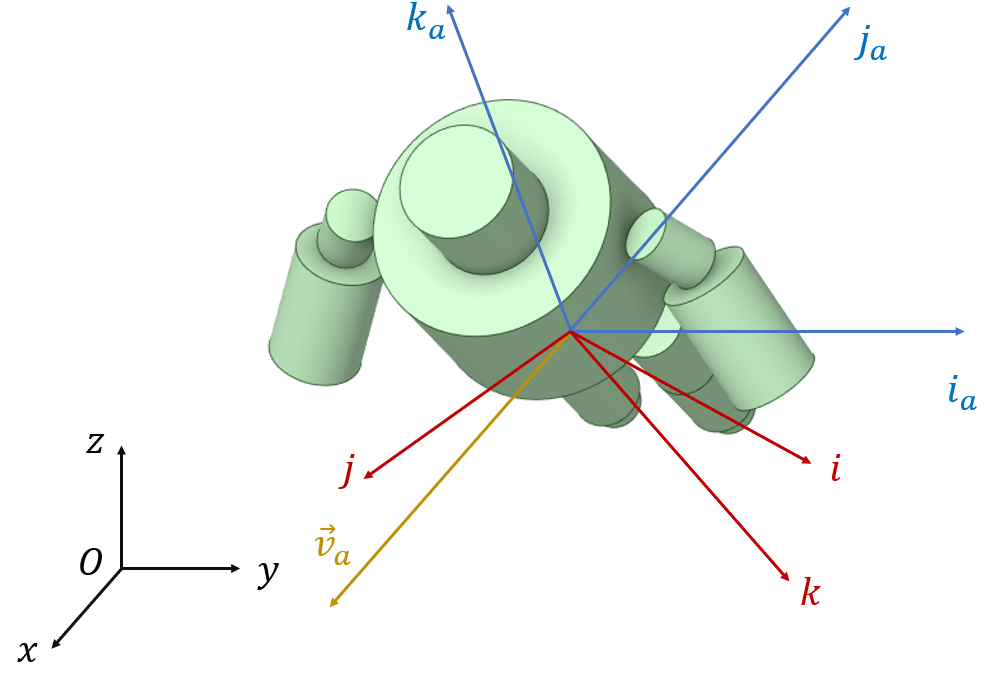}
	\caption{CFD Model with the relative velocity frame.}
	\label{figureRelativeVelocityFrameCFD}
\end{figure}

For the sole purpose of CFD analysis, let us introduce the frame $\mathcal{A}=\{{}^{\mathcal{I}}o_{\mathcal{G}};\boldsymbol{i}_a,\boldsymbol{j}_a,\boldsymbol{k}_a\}$ as the \emph{relative velocity} frame. It is defined for CFD simulations only, where the relative velocity is always different from zero, namely $|\boldsymbol{v}_a|\neq 0$. 
The $\boldsymbol{j}_a$ axis therefore aligns to the opposite direction of $\boldsymbol{v}_a$ (see also Fig. \ref{figureRelativeVelocityFrameCFD}).

Then, we decompose the aerodynamic force w.r.t. the relative velocity frame $\mathcal{A}$ as:
\begin{equation}
	\label{eq:af_total_control}
	\boldsymbol{{F}_a}=\boldsymbol{{F}_D}+\boldsymbol{{F}_S}+\boldsymbol{{F}_N} ,
\end{equation}
where:
\begin{itemize}
	\item $\boldsymbol{{F}_D}$, the drag force, aligns to $\boldsymbol{j}_a$ axis;
	\item $\boldsymbol{{F}_S}$, the defined sideforce, aligns to $\boldsymbol{i}_a$ axis;
	\item $\boldsymbol{{F}_N}$, the defined normal force, aligns to $\boldsymbol{k}_a$ axis.
\end{itemize}
The intensity of the steady aerodynamic force varies approximately with $|\boldsymbol{v}_a|^2$, hence there exist three dimensionless functions $C_D$, $C_S$, $C_N$ depending on the Reynolds number $Re$, the Mach number $\mathcal{M}$ and $(\alpha, \beta)$, such that \cite{c3}\cite{c5}:
\begin{subequations}
	\begin{align}
		\boldsymbol{{F}_D} & :=\mathcal{K}_a|\boldsymbol{v}_a|^2C_D(Re, \mathcal{M}, \alpha, \beta)\boldsymbol{j}_a,\\
		\boldsymbol{{F}_S} & :=\mathcal{K}_a|\boldsymbol{v}_a|^2C_S(Re, \mathcal{M}, \alpha, \beta)\boldsymbol{i}_a,\\
		\boldsymbol{{F}_N} & :=\mathcal{K}_a|\boldsymbol{v}_a|^2C_N(Re, \mathcal{M}, \alpha, \beta)\boldsymbol{k}_a,
	\end{align}
	\label{eq:af_coe}
\end{subequations}
\noindent with the angles $(\alpha, \beta)$ characterizing the relative orientation of $\boldsymbol{v}_a$ w.r.t. the body frame $\mathcal{D}$, chosen as in  Fig.~\ref{fig:notation}. So, $\alpha \in [0,\pi]$ is defined as the angle between $\boldsymbol{v}_a$ and the negative $\boldsymbol{k}$ axis of the body frame $\mathcal{D}$, while $\beta \in [0,2\pi]$ is defined as the angle between the projection of $\boldsymbol{v}_a$ on plane $(\boldsymbol{i},\boldsymbol{j})$ and the positive $\boldsymbol{i}$ axis of frame $\mathcal{D}$ (see also \cite{c3}). The constant $\mathcal{K}_a:=\frac{A_{ref}\rho}{2}$ with $\rho$ the air density and $A_{ref}$ the reference frontal area, moreover, $A_{ref}$ has been defined to obtain $\mathcal{K}_a = 1$ to simplify the calculation. Furthermore, we hereafter assume that the dependency upon the Mach and Reynolds numbers is negligible, and thus omitting these dependencies in the arguments of the aerodynamic coefficients. 


\subsection{CFD results and simplified aerodynamic model}\label{sec:aeroModel}

Significant robot flight configurations, corresponding to precise values of $\alpha$ and $\beta$, were simulated using CFD tools. In total 45 simulations were run, with $ \alpha \in (\ang{15},\ \ang{30},\ \ang{45},\ \ang{60},\ \ang{90},\ \ang{120},\ \ang{150},\ \ang{160},\ \ang{180})$ and $ \beta \in (\ang{90},\ \ang{135},\ \ang{180},\ \ang{225},\ \ang{270})$ which supplied the necessary data for proceeding with the modeling of aerodynamic characteristics on iRonCub. Settings of CFD simulations are schematically reported in Table \ref{table:cfd}.

\begin{table}[t]
	\caption{CFD simulation settings}
	\label{table:cfd}
	\centering
	\begin{tabular}{c|l}
		\toprule
		\multirow{6}{*}{\textit{Solver Settings}} & Pressure--Velocity Coupling for \emph{RANS} \\ & (Reynolds Averaged Navier-Stokes) \\ & Steady-State Simulation\\ & SST $k - \omega$ for Turbulence modelling \\ & No Compressibility Effects \\ & (i.e. $M\ll1$, $\rho = \text{const}$)\\
		\midrule
		\multirow{2}{*}{\textit{Constant Values}} & Air density: $\rho = \SI{1.225}{kg/m^3}$ \\ & Reference pressure: $p_{ref} = \SI{1}{atm}$\\
		\midrule
		\multirow{3}{*}{\textit{Boundary Conditions}} & Inlet velocity: $v = \SI{7.5}{m/s}$ \\ & Outlet pressure: $p_{rel} = \SI{0}{Pa}$\\ & Walls: \textbf{No Slip} Condition\\
		\bottomrule
	\end{tabular}
\end{table}

\looseness=-1

As introduced in Sec. \ref{sec:cfd}, the three dimensionless functions $C_D$, $C_S$, $C_N$ are the aerodynamic characteristics of the body, i.e. the defined aerodynamic coefficients (AC): drag coefficient, sideforce coefficient and normal force coefficient. In this section, the analytical model of AC is identified based on the CFD results from Sec. \ref{sec:cfd}. To proceed with identification, the following assumptions are made:

\begin{enumerate}
	\item the aerodynamic forces act at the robot's CoM (coincides with CoP) and therefore the aerodynamic moments ${M}_a$ are equal to zero;
	\item from CFD results, the sideforce effect is neglected due to its small magnitude compared with the other two aerodynamic force components, and with other forces acting on the robot;
	\item the AC dependence from Mach number $\mathcal{M}$ is considered negligible since the relative velocity intensity of the robot is much smaller than the sound speed (i.e. the density can be assumed as constant);
	\item the AC dependence from Reynolds number $Re$ is not considered since the AC evaluated from CFD for different values of relative velocity doesn't show a significant dependence from it;
	\item the effects of rotational and unsteady motions of the vehicle on its surrounding airflow are negligible~\cite{c5}.
\end{enumerate}

As a consequence, the AC depend only on the flight configuration $(\alpha,\beta)$ and can be written as~\cite{c3}:
\begin{equation*}
	C_D(\cdot) =C_D(\alpha, \beta), \quad  C_N(\cdot) =C_N(\alpha, \beta) .
\end{equation*}
Considering that the aerodynamic coefficients are naturally $2\pi$-periodic \cite{c12}, $\sin{(\cdot)}$ functions are used to design the mathematical models of AC: 
\begin{equation}
	\resizebox{.88 \columnwidth}{!}{$ 
		C_D(\alpha,\beta)= c_0 + c_1 \sin{^2 \alpha} \sin{^2\beta} + c_2 \sin{^2\alpha}  + c_3 \sin{^2\beta}
		$},\label{eq:drag}
\end{equation}
\begin{equation}
	C_N(\alpha,\beta)= d_0 + d_1 \sin{^2\alpha} \sin{(2\alpha)} \sin{^2\beta} .
	\label{eq:normal}
\end{equation}

\begin{figure}[t]
	\centering
	\subfigure[\label{fig:drag model}]{\includegraphics[width=0.97\columnwidth]{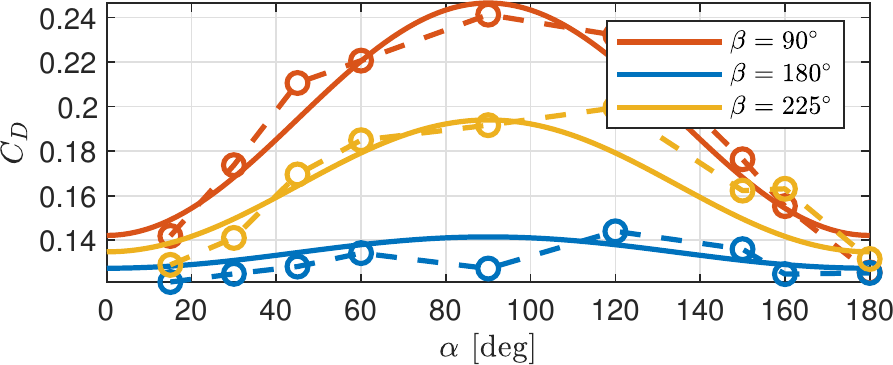}}\\
	\subfigure[\label{fig:normal model}]{\includegraphics[width=0.97\columnwidth]{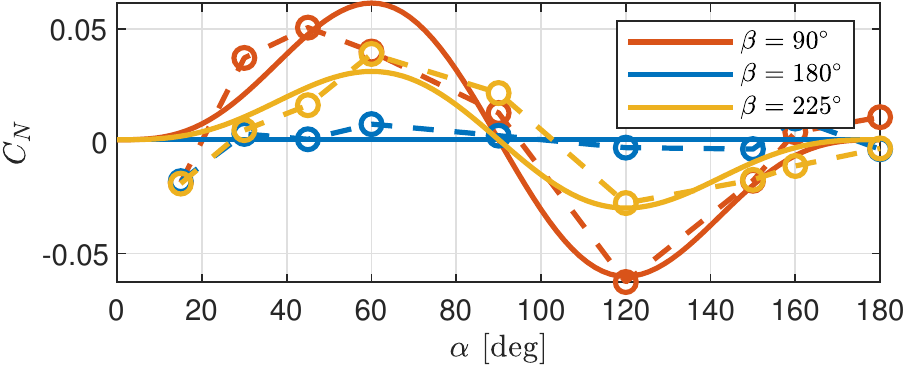}} \\
	\caption{AC models identified from CFD simulations data. Dashed lines represent the collected CFD data, solid lines represent the identified models.}
	\label{fig:subfig}
\end{figure}

The coefficients characterizing the expressions of $C_D(\cdot)$ and $C_N(\cdot)$:
\begin{itemize}
	\item $c_0=0.1274$, $c_1=0.0903$, $c_2=0.0141$, $c_3=0.0147$;
	\item $d_0=0.0007$, $d_1=0.0938$,
\end{itemize}
\noindent
are obtained using linear regression. \\
In fluid dynamics, drag force is a type of friction that acts in the opposite direction of the relative motion of any object moving in a fluid. Therefore the estimated drag coefficient $C_D$ should remain positive in the whole range of $\alpha$ and $\beta$. This property is verified by the identified model in Eq.~\eqref{eq:drag}, as Fig.~\ref{fig:drag model} shows.

As previously mentioned, the zero sideforce assumption allows the identification of the lift force with the normal force, as demonstrated in~\cite{c3} for axisymmetric bodies, since the normal component of the aerodynamic force lies on the $\boldsymbol{v}_a-\boldsymbol{k}_a$ plane. The aerodynamic forces acting on the robot CoM rewrite as:

\begin{equation}
	\label{eq:aerodyModelSymmetricBodies}
	\begin{array}{lcl}
		\boldsymbol{F_L} & = & k_a|\boldsymbol{v}_a|C_L(\alpha,\beta)\boldsymbol{r}(\beta) \times \boldsymbol{v}_a, \\
		\boldsymbol{F_D} & = & -k_a |\boldsymbol{v}_a|C_D(\alpha,\beta)\boldsymbol{v}_a, \\
	\end{array}
\end{equation}
where  $\label{eq:aerodynamicsAuxiliaryVector}
\boldsymbol{r}(\beta)  =  -\sin(\beta) \boldsymbol{i}+\cos(\beta) \boldsymbol{j}.$

\section{Control Design}
\label{sec:control}


\subsection{Flight Control Without Aerodynamics Awareness}
\label{subsec:Flight-Control-no-aerodyn-forces}
The iRonCub flight controller~\cite{c1, c2} is a \emph{task-based} control strategy, composed of several control objectives (tasks), with different priorities. The so-called \emph{primary task}, i.e. the task with the higher priority, is the stabilization of the robot's \emph{centroidal momentum}\footnote{The centroidal momentum is the robot total momentum expressed w.r.t. the frame $\mathcal{G}[\mathcal{I}]$  defined in Sec. \ref{background}.} dynamics, which equals the summation of all external forces acting on the robot:
\begin{equation}
	\label{eq:hdot}
	{}^{\mathcal{G}[\mathcal{I}]}\dot{\boldsymbol{h}}=\boldsymbol{A}(q)T+m\bar{\boldsymbol{g}}=f_h(q,T),
\end{equation}

where $m$ is the robot's total mass, $\boldsymbol{A}(q) \in \mathbb{R}^{6 \times m}$ is a proper matrix mapping the thrust vector $T$ in the momentum equation. For the momentum control design, we assume that the robot joints velocities $\dot{s}$ and the thrust rate of change $\dot{T}$ can be chosen at will, and then can be considered as our control inputs. Therefore, we define $u := (\dot{T}, \dot{s})$. To make the control input $u$ appear in the momentum equation \eqref{eq:hdot}, we increase the relative degree of the output and write the momentum acceleration:
\begin{equation}
	\label{eq:hddot}
	{}^{\mathcal{G}[\mathcal{I}]}\ddot{\boldsymbol{h}}=\dot{\boldsymbol{A}}(q)T+\boldsymbol{A}(q)\dot{T} = f_h(q, T, v, \dot{T}).
\end{equation}
As demonstrated in our previous works \cite{c1,c2}, it is possible to find a smooth control input $u^*$ that renders the closed-loop equilibrium point $(\dot{\tilde{\boldsymbol{h}}}, \tilde{\boldsymbol{h}}, \boldsymbol{I}) = (\boldsymbol{0},\boldsymbol{0},\boldsymbol{0})$ globally asymptotically stable, where $\tilde{\boldsymbol{h}} = \boldsymbol{h}-\boldsymbol{h}^d$, $\boldsymbol{h}^d$ is the reference momentum trajectory and $\boldsymbol{I} := \int_0^t \tilde{\boldsymbol{h}}$. 

We also define a \emph{postural task} to resolve the redundancy in the control input $u$ and maintain the robot configuration close to a desired shape:
\begin{equation}
	\label{eq:postural}
	\dot{s}^* = -K_P^s (s-s^d),
\end{equation}
with $s^d$ the desired robot posture and $K_P^s \in \mathbb{R}^{n \times n}$ a positive gains matrix. The two tasks are then achieved by means of Quadratic Programming (QP) optimization framework:
\begin{IEEEeqnarray}{LLL}
	\IEEEyesnumber
	\label{eq:optimization_QP}
	u^{**} &=& \text{argmin}_{u} (w_1|u-u^*|^2 + w_2|\dot{s}-\dot{s}^*|^2) \\
	& s.t. & \qquad u_{min} \leq u \leq u_{max}, \nonumber
\end{IEEEeqnarray}
where \emph{soft} task priorities are assigned to the two tasks via proper weights $w_1, w_2 >0$. Framing the problem as an optimization procedure allows to include input boundaries $u_{min}, u_{max}$ in the control design. It is also possible to include the integral of input boundaries (namely, the joint position and thrust limits) in the QP design. For more details on these aspects of the flight control design, which are not relevant for this paper, we refer to previous works \cite{c1,c2}.

\subsection{Flight Control with Aerodynamics Awareness}

The flight controller discussed in \ref{subsec:Flight-Control-no-aerodyn-forces} does not consider the effect of aerodynamic forces. This missing information may impair the performances of the control algorithm when aerodynamic forces are not negligible. To overcome this limitation, we implemented two modifications of the iRonCub flight controller: \emph{feedback linearization} and \emph{gain scheduling}.
To implement our control strategy, we considered the following \textbf{Assumption 2}: we assume that $\dot{\boldsymbol{F}}_a = 0$ in the calculation of the momentum acceleration.

\subsubsection*{Addition of Aerodynamics Forces}
First, we include the estimated aerodynamic forces, obtained by writing Eq. \eqref{eq:af_total_control} with the aerodynamic coefficients estimated in Sec. \ref{sec:aeroModel}, in the robot equations of motion Eq. \eqref{eq_motion} and momentum dynamics Eq. \eqref{eq:hdot}, which renders:
\begin{equation}
	\label{eq_motion_af}
	M(q)\dot{v}+C(q,v)v+G(q)=
	\begin{bmatrix}
		0_6\\
		\tau
	\end{bmatrix}
	+f(q,T)+J_{CoM}^{\top}F_a,
\end{equation}
where $J_{CoM} \in \mathbb{R}^{3 \times n+6}$ is the CoM Jacobian, while the momentum rate of change is now given by:
\begin{equation}
	\label{eq:hdot_af}
	{}^{\mathcal{G}[\mathcal{I}]}\dot{\boldsymbol{h}}=\boldsymbol{A}(q)T+m\bar{\boldsymbol{g}}+
	\begin{bmatrix}
		\boldsymbol{F_a}\\
		0_3
	\end{bmatrix}.
\end{equation}

\subsubsection*{Feedback Linearization Control}

We assume that the relative velocity ${}^{\mathcal{I}}\boldsymbol{v}_a$ can be measured, for example by employing dedicated sensors such as Pitot tubes applied on the robot, allowing the computation of the aerodynamic force $F_a$. A feedback linearization controller then substitutes Eq.~\eqref{eq:hdot} with Eq.~\eqref{eq:hdot_af} in the calculation of the baseline control input $u^*$, which depends on the momentum rate of change $\dot{h}$. In this way we can cancel out the effects of the aerodynamic forces on the desired closed-loop momentum acceleration.

\subsubsection*{Implementation of Gain Scheduling}

if it is not possible to compute the aerodynamic forces with enough accuracy, gain scheduling is employed as a strategy to react to wind gusts. In particular, we assume that a wind gust is detected when the measured center of mass position error overcomes a certain threshold.

The desired closed-loop momentum acceleration has the following shape, which is obtained by applying the baseline controller $u^*$ Eq. \eqref{eq:optimization_QP}:
\begin{equation}
	\label{eq:hddot_des}
	\ddot{\boldsymbol{h}} = \ddot{\boldsymbol{h}}^d -K_D\dot{\tilde{\boldsymbol{h}}} -K_P\tilde{{\boldsymbol{h}}} -K_I \boldsymbol{I}.
\end{equation}
When a wind gust is detected, gain scheduling smoothly increases the value of the gain $K_I$ and $K_P$, to guarantee a stronger response of the controller to the associated errors.

\section{Simulation results}
\label{simulation}

\subsection{Simulation Environment}


The robot dynamics is simulated by numerically integrating the equations of motion with aerodynamic forces Eq.~\eqref{eq_motion_af} in a custom-made simulator implemented in MATLAB-Simulink. The controller is entirely designed in Simulink and runs at a frequency of $\SI{100}{Hz}$. A fast, inner torque control loop is designed to stabilize the joints velocities $\dot{s}$ towards the reference values generated with Eq. \eqref{eq:optimization_QP}.\looseness=-1

\subsection{Flight Envelope Design}

\begin{figure}[t]
	\centering
	\subfigure{\label{fig:high_speed_wind}
		\includegraphics[width=0.97\columnwidth]{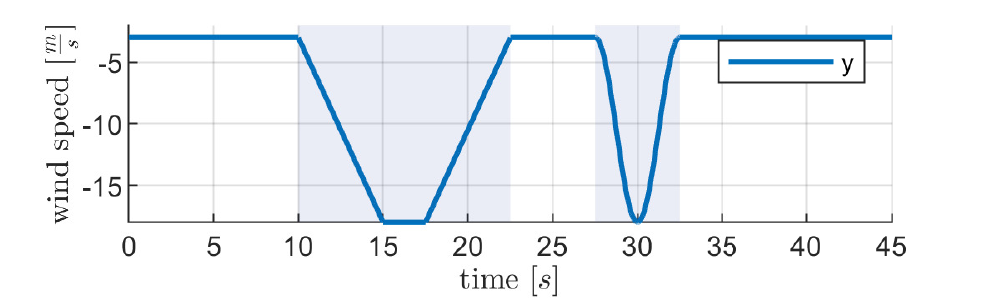}}
	\subfigure{\label{fig:high_speed_angMom}
		\includegraphics[width=0.97\columnwidth]{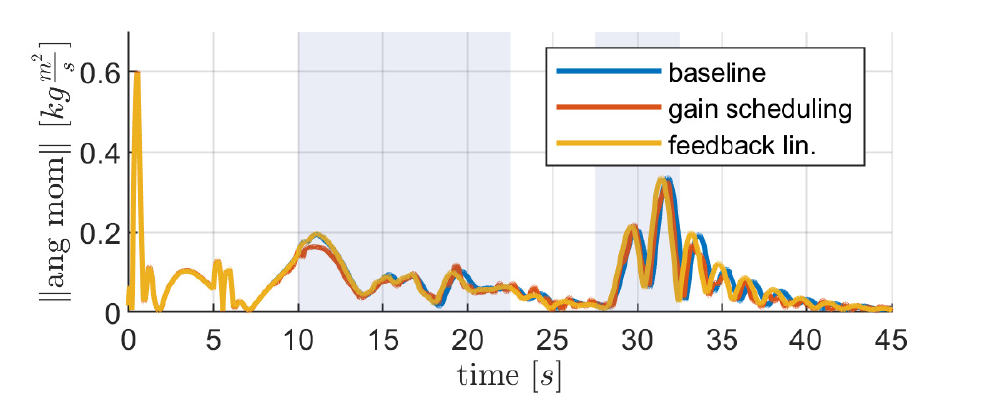}}
	\subfigure{\label{fig:high_speed_linMom}
		\includegraphics[width=0.97\columnwidth]{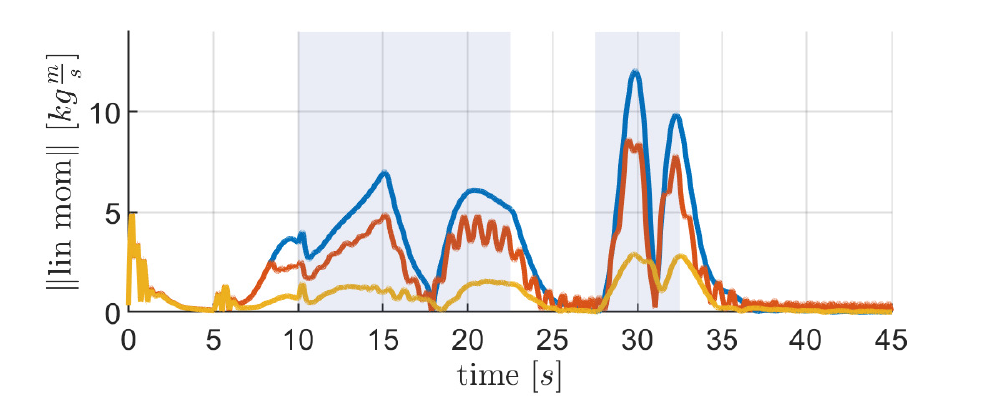}}
	\subfigure{\label{fig:high_speed_com}
		\includegraphics[width=0.97\columnwidth]{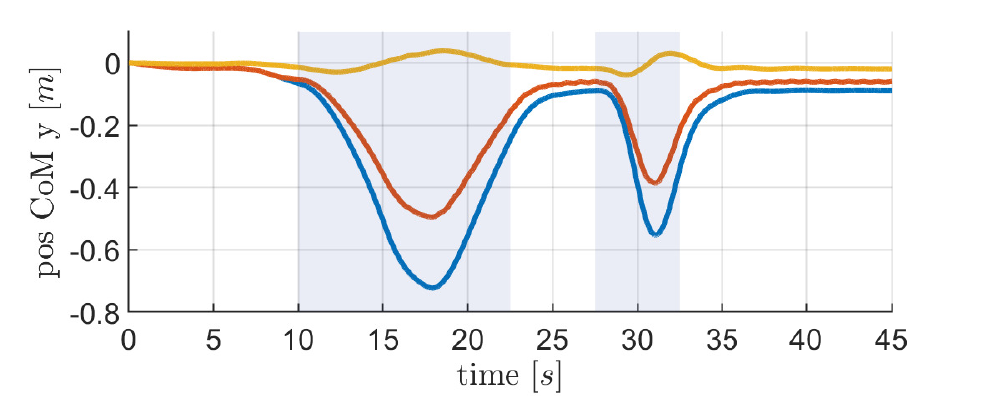}}        
	\caption[]{High speed flight scenario. From top to bottom: a) wind profile; b) angular momentum error norm; c) linear momentum error norm; d) center of mass position error in the wind direction.}
	\label{fig:high_speed_plots}
\end{figure}

We designed a flight envelope composed as follows:

\begin{enumerate}
	\item \textit{Scenario 1}: the robot hovers in vertical position, while subject to a wind in the frontal direction that follows the profile depicted in Fig. \ref{fig:hovering_wind}. 
	\item \textit{Scenario 2}: the robot hovers and climbs to gain altitude, then moves forward at around $\SI{11}{m/s}$, in presence of a lateral wind with the profile of Fig. \ref{fig:high_speed_wind}.
\end{enumerate}
The wind profiles are composed of a constant wind of $\SI{3}{m/s}$, and wind gusts of different shapes with magnitude of $\SI{10}{m/s}$ and $\SI{15}{m/s}$ \cite{aero,wind,KAI2019108491}. 



\subsection{Control Results for the Hovering Scenario}

\begin{figure}[t]
	\centering
	\subfigure{\label{fig:hovering_wind} \includegraphics[width=0.97\columnwidth]{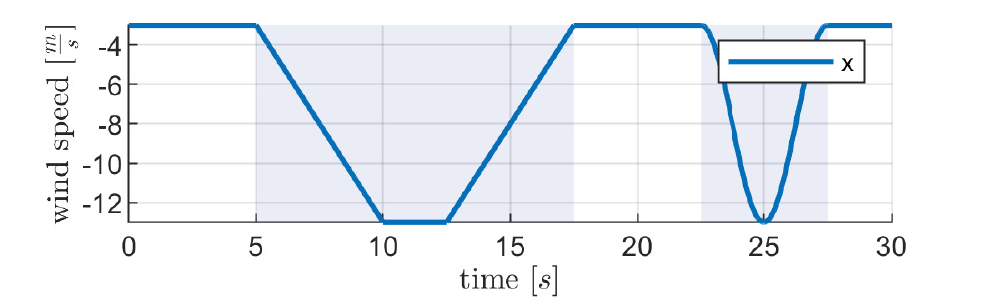}}
	\subfigure{\label{fig:hovering_angMom}
		\includegraphics[width=0.97\columnwidth]{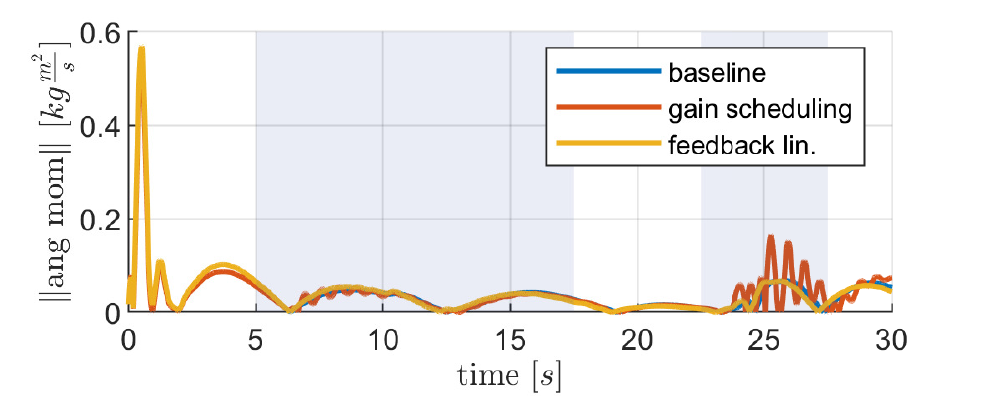}}
	\subfigure{\label{fig:hovering_linMom}
		\includegraphics[width=0.97\columnwidth]{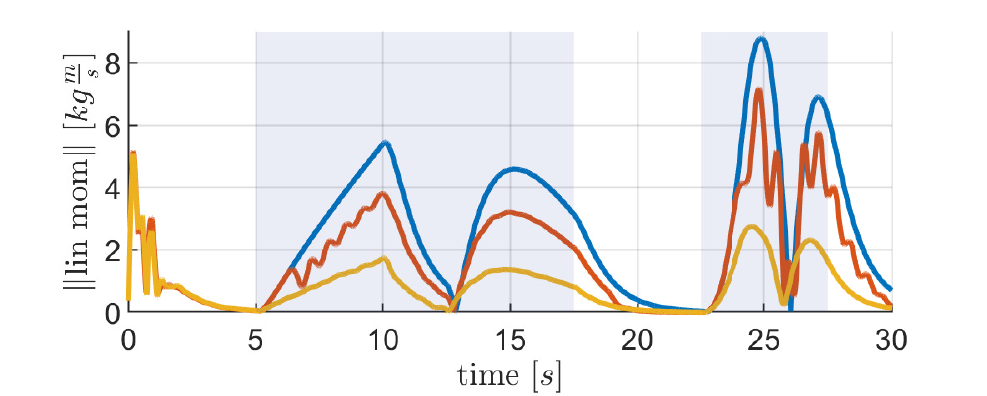}}
	\vspace{-0.1cm}
	\subfigure{\label{fig:hovering_com}\includegraphics[width=0.97\columnwidth]{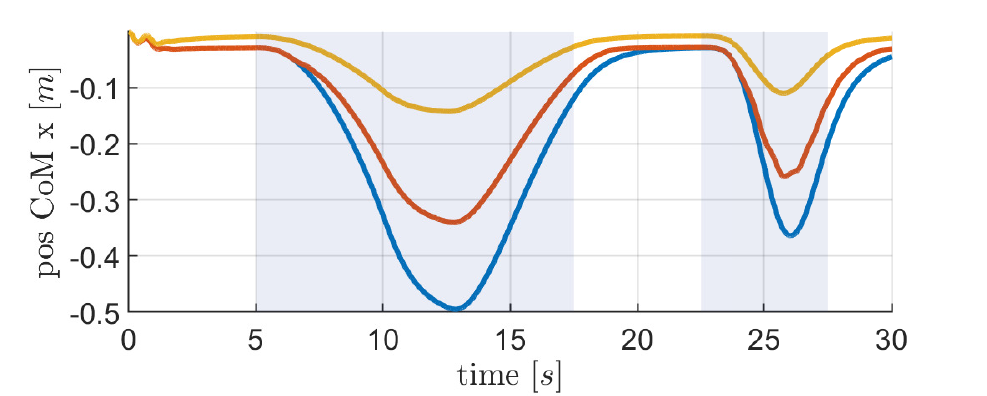}}        
	\caption[]{Hovering scenario. From top to bottom: a) wind profile; b) angular momentum error norm; c) linear momentum error norm; d) center of mass position error in the wind direction.}
	\label{fig:hovering_plots}
\end{figure}

We evaluate the performances of three control strategies: the \emph{baseline} controller of \cite{c1}\cite{c2}, the \emph{feedback linearization} control designed in Sec. \ref{sec:control}, and a third controller that uses \emph{gain scheduling} as described in Sec. \ref{sec:control}. To test the robustness of feedback linearization controller, the aerodynamic force $F_a$ in the control model is computed assuming random noise of $\SI{5}{\%}$ and a calibration error of $\SI{10}{\%}$ on both $\alpha$ and $\beta$ angles.

Simulation results for the hovering scenario are presented in Fig.~\ref{fig:hovering_plots}, where we focused on the linear and angular momentum error norm and center of mass position error. With baseline controller, the center of mass error along the wind direction has peaks of $\SI{50}{cm}$ and $\SI{35}{cm}$ in correspondence of the wind gusts (shaded area), which are reduced by $\SI{33}{\%}$ when the gain scheduling control is used and by $\SI{71}{\%}$ for the feedback linearization control. The linear momentum error also reduces accordingly, while the angular momentum error remains comparable in all three simulations.

\subsection{Control Results for High Speed Flight}
\label{subsec:ctrl-results-aerodyn}

The three controllers are then tested during high speed flight. Fig.~\ref{fig:high_speed_plots} compare their performances in this second scenario: the baseline controller has the worst performances, with peaks errors on the center of mass position of $\SI{75}{cm}$ and $\SI{50}{cm}$. The gain scheduling controller is able to reduce the center of mass peaks error by $\SI{30}{\%}$, while the feedback linearization further reduces the peaks by $\SI{95}{\%}$. Linear and angular momentum errors also behave in accordance with the results already achieved for Scenario 1.

\section{Conclusions}
\label{conclusion}


We proposed a strategy to model and control a flying multibody robot in presence of non-negligible aerodynamic effects. The model of aerodynamic forces is identified based on the data provided by CFD simulations. These forces are then used to design appropriate controllers for the robot flying in a given flight envelope. 
Two techniques, \emph{gain-scheduling} and \emph{feedback linearization}, are applied to improve the flight controller performances in presence of aerodynamic effects, and all controllers were tested successfully in simulations.

Nevertheless, the proposed framework requires several assumptions, both for performing the CFD analysis and in the controller design. Future work may consider to relax these assumptions, for example by replacing the robot model for CFD analysis with a more accurate one, closer to the actual shape of iRonCub. Another future work will involve running CFD simulations with the robot in different joints configurations.
Finally, further effort in control design is required to improve the robustness of the controller for different flight envelopes.\looseness=-1




\end{document}